\documentclass[conference]{IEEEtran}
\IEEEoverridecommandlockouts
\usepackage{cite}
\usepackage{amsmath,amssymb}
\usepackage{graphicx}
\usepackage{textcomp}
\usepackage{xcolor}
\usepackage{times}
\usepackage{soul}
\usepackage{url}
\usepackage[hidelinks]{hyperref}
\usepackage[utf8]{inputenc}
\usepackage[small]{caption}
\usepackage{graphicx}
\usepackage{amsthm}
\usepackage{booktabs}
\usepackage[linesnumbered,ruled,vlined]{algorithm2e}
\usepackage[switch]{lineno}
\usepackage{diagbox}
\usepackage{multirow}
\usepackage{CJKutf8}
\usepackage{subfigure}
\usepackage{bm}
\usepackage{caption}

\DeclareMathAlphabet\mathbfcal{OMS}{cmsy}{b}{n} 
 
\newcommand{\mat}[1]{\mathbf{#1}}
\usepackage{tabularx} 

\usepackage[normalem]{ulem}
\useunder{\uline}{\ul}{}
\usepackage{xcolor}

\DeclareMathAlphabet\mathbfcal{OMS}{cmsy}{b}{n}

\def\BibTeX{{\rm B\kern-.05em{\sc i\kern-.025em b}\kern-.08em
    T\kern-.1667em\lower.7ex\hbox{E}\kern-.125emX}}
    
\begin{document}

\title{Spatial-Temporal Large Language Model for Traffic Prediction}

\author{\IEEEauthorblockN{Chenxi Liu$^{1}$, Sun Yang$^{2}$, Qianxiong Xu$^{1}$, Zhishuai Li$^{3}$, Cheng Long$^{1, *}$, Ziyue Li$^{4, *}$\thanks{* Corresponding authors.}, Rui Zhao$^{3}$}

\IEEEauthorblockA{
$^{1}$\textit{S-Lab, Nanyang Technological University, Singapore}\\
$^{2}$\textit{School of Software and Microelectronics, Peking University, China}\\
$^{3}$\textit{SenseTime Research, China}\\
$^{4}$\textit{Information System Department, University of Cologne, Germany}\\
{\{chenxi.liu, qianxiong.xu, c.long\}@ntu.edu.sg, 2201210484@stu.pku.edu.cn}\\
{lizhishuai@sensetime.com, zlibn@wiso.uni-koeln.de, zhaorui@sensetime.com}
}
}

\maketitle

\begin{abstract}
Traffic prediction, an essential component for intelligent transportation systems, endeavours to use historical data to foresee future traffic features at specific locations. Although existing traffic prediction models often emphasize developing complex neural network structures, their accuracy has not improved. Recently, large language models have shown outstanding capabilities in time series analysis. Differing from existing models, LLMs progress mainly through parameter expansion and extensive pretraining while maintaining their fundamental structures. Motivated by these developments, we propose a Spatial-Temporal Large Language Model (ST-LLM) for traffic prediction. In the ST-LLM, we define timesteps at each location as tokens and design a spatial-temporal embedding to learn the spatial location and global temporal patterns of these tokens. Additionally, we integrate these embeddings by a fusion convolution to each token for a unified spatial-temporal representation. Furthermore, we innovate a partially frozen attention strategy to adapt the LLM to capture global spatial-temporal dependencies for traffic prediction. Comprehensive experiments on real traffic datasets offer evidence that ST-LLM is a powerful spatial-temporal learner that outperforms state-of-the-art models. Notably, the ST-LLM also exhibits robust performance in both few-shot and zero-shot prediction scenarios. The code is publicly available at https://github.com/ChenxiLiu-HNU/ST-LLM.
\end{abstract}


\section{Introduction}
\label{sec:intro}
Traffic prediction, which aims to predict future traffic features like traffic flow at specific locations using historical data, is a crucial component for intelligent transportation systems~\cite{DBLP:conf/aaai/JiangHZW23,chang2023tensor,DBLP:journals/tkdd/LiFYJYSJL23,gong2023empowering,liu2021understanding,DBLP:journals/tits/XiaoX0HLZ022,DBLP:conf/cikm/XuRLYZ22}. This prediction is instrumental in optimizing traffic management~\cite{stpicde24,zhou2024crest,chenxi2021study} and scheduling public transportation~\cite{10328393,10.1145/3340531.3412054,DBLP:conf/hpcc/ChenWL20}. For instance, accurately predicting bike flow benefits the transportation department in optimizing bike management. Similarly, forecasting taxi flow is vital for taxi companies, as it enables them to efficiently allocate and schedule vehicles to satisfy expected demand~\cite{liu2024spatial,DBLP:journals/isci/JinLXSLH22,DBLP:conf/mdm/HeZCLLLS23}.

The evolution of traffic prediction has seen a shift from traditional time series models to deep learning techniques~\cite{DBLP:journals/pacmmod/0002Z0KGJ23,yuan2018hetero,kumar2015short,liu2024icde}. Initially, time series models such as the Autoregressive Integrated Moving Average and Kalman Filter were adapted for their fit with time series data. However, these models are not good at capturing the spatial-temporal dependencies within traffic data, leading to deep learning solutions using convolutional neural networks (CNNs) for spatial and recurrent neural networks (RNNs) for temporal dependencies~\cite{yin2021deep,shen2018stepdeep,yuan2018hetero,DBLP:journals/www/CaiWCLX24}. Despite these advancements, the non-Euclidean spatial structure and the complex periodicity of traffic data present challenges for CNNs and RNNs in capturing spatial and temporal dependencies well.

Graph convolutional network (GCN) based models gained popularity for their ability to model local spatial dependencies~\cite{DBLP:journals/tkdd/LiFYJYSJL23,bai2020adaptive,DBLP:conf/ijcai/WuPLJZ19,li2018dcrnn_traffic,10.5555/3304222.3304273,miao2023task,zhong2023personalized,xu2023kits}. However, these models often encountered over-smoothing issues, which limits their ability to capture global spatial patterns. This shortcoming prompted a shift to attention-based models, which effectively model dynamic spatial correlations without depending on an adjacency matrix~\cite{DBLP:conf/aaai/JiangHZW23,9346058,DBLP:conf/aaai/LinLZCY20}. These attention-based approaches have since emerged as a leading trend, offering a superior ability to handle spatial-temporal dependencies in traffic prediction~\cite{zheng2020gman,DBLP:conf/aaai/GuoLFSW19}. Nevertheless, with this evolution, the structures of existing traffic prediction models have become progressively complex. 

Foundation models, including large language models (LLMs), have advancements in fields such as computer vision~\cite{DBLP:conf/cvpr/KoCCORK23,DBLP:conf/cvpr/YuWFZL23} and natural language processing~\cite{DBLP:conf/acl/Ramezani023,DBLP:conf/acl/MaynezAG23}. More recently, LLMs also have shown superb performance on time series analysis~\cite{xue2022leveraging,xue2023promptcast,cao2023tempo,jin2023time}. Compared with the complex designs of existing predictive models, LLMs primarily evolve by expanding parameters and pretraining while maintaining their foundational model structure. Existing LLM-based prediction methods focus on the temporal aspect of data in the traffic prediction tasks~\cite{zhou2023onefitsall,jin2023time,cao2023tempo} and often overlook the spatial aspect. However, in traffic prediction, the spatial variables are strongly correlated and the spatial dimension also proves to be important~\cite{lablack2023spatio,10.1145/3589132.3625614}. For example, a common setting is to use traffic data from the previous \emph{twelve} timesteps to predict traffic for the next \emph{twelve} timesteps at \emph{hundreds} of spatial locations~\cite{li2018dcrnn_traffic} - in this case, more spatial data than temporal data can be leveraged. In our study, 
we define the timesteps of a spatial location as a token and model the global temporal dependencies across all these tokens to emphasize the spatial aspects.

Moreover, LLMs are notable for their ability to transfer knowledge across domains~\cite{gruver2023llmtime}, such as the pretrained transformer (FPT) LLM~\cite{zhou2023onefitsall}. While the FPT LLM is effective in time series analysis tasks, it shows less optimal performance in long-term prediction tasks like traffic prediction. The possible reason is that FPT struggles to bridge the domain gap between language and traffic data.
To fill this gap, we propose a partially frozen attention (PFA) LLM specifically designed to enhance traffic prediction accuracy. By partially freezing the multi-head attention layers, the LLM can adapt to traffic prediction while preserving the foundational knowledge acquired during pretraining.

In summary, we propose a novel Spatial-Temporal Large Language Model (ST-LLM) for traffic prediction. Within the ST-LLM framework, we define timesteps at a location as a token. These tokens transform a specialized spatial-temporal embedding layer, which is designed to emphasize spatial locations and global temporal patterns. Furthermore, we fuse the spatial-temporal embeddings of each token for a unified representation. Following this, we introduce the partially frozen attention LLM, a novel strategy tailored for LLMs to capture the global spatial-temporal dependencies in traffic prediction effectively. Extensive experiments on real-world traffic datasets have validated the efficacy of ST-LLM. The key contributions of this paper are summarized as follows:

\begin{itemize}

    \item We propose a Spatial-Temporal Large Language Model (ST-LLM) for traffic prediction, which defines timesteps at a location as a token and embeds each token by a spatial-temporal embedding layer. We fuse the spatial-temporal embeddings of these tokens uniformly and adapt the LLMs to capture global spatial-temporal dependencies.
    
    \item A novel strategy within the LLM, named partially frozen attention, is proposed to enhance the model in traffic prediction. By partially freezing the multi-head attention, the ST-LLM is adapted to capture global spatial-temporal dependencies between tokens for different traffic prediction tasks.

    \item Extensive experiments are conducted on real traffic datasets to show the superior performance achieved by our ST-LLM across various settings. Moreover, the few-shot and zero-shot prediction results highlight the ST-LLM's capability for intra-domain and inter-domain knowledge transfer.
    
\end{itemize}

The remainder of this paper is as follows. Section~\ref{sec: related work} discusses related work about LLMs for time series analysis and traffic prediction. Section~\ref{sec: definition} introduces the problem definition. Section~\ref{sec: methodology} details the ST-LLM, followed by the experiments in Section~\ref{sec: experiments}. Section~\ref{sec: conclusion} concludes the paper.
	
\section{Related Work}
\label{sec: related work}

In this section, we review the related work from two perspectives, large language models for time series analysis and traffic prediction.

\subsection{Large Language Models for Time Series Analysis}

Recently large language models (LLMs) have shown superb performance on time series analysis tasks~\cite{jin2023large}, such as prediction~\cite{cao2023tempo}, classification~\cite{jin2023time}, anomaly detection~\cite{zhou2023onefitsall}, imputation~\cite{chen2023gatgpt}, few-shot learning~\cite{sun2023test}, and zero-shot learning~\cite{gruver2023llmtime}.
For instance, TEMPO-GPT combined prompt engineering and seasonal trend decomposition within its generative pretrained transformer (GPT) structure~\cite{cao2023tempo}. This integration enabled the model to recall pertinent knowledge from historical data, matching time series inputs with distinct temporal semantic elements.
TIME-LLM reprogrammed an LLM for time series forecasting, and the backbone language model remained intact~\cite{jin2023time}. The authors reprogrammed the input time series with text prototypes before feeding it into the frozen LLM to align the text and time series modalities.
OFA employed a frozen GPT2 model across various key tasks in time series analysis~\cite{zhou2023onefitsall}, the authors concluded that the LLM performed better on tasks of time series, such as imputation, classification, anomaly detection, and few-shot learning. 
TEST executed time series forecasting and classification tasks~\cite{sun2023test} and generated the similarity-based, instance-wise, feature-wise, and text-prototype-aligned embedding for time series tokens. LLMTIME leveraged pretrained LLMs for continuous time series forecasting by representing numbers in text format and generating possible extrapolations through text completions~\cite{gruver2023llmtime}. The above model only models the temporal dimension of the data and ignores the spatial dimension. GATGPT integrates the graph attention network and GPT for spatial-temporal imputation, and the graph attention mechanism boosts the LLM's capability to grasp spatial dependencies~\cite{chen2023gatgpt}. However, it directly overlooks the temporal representation. As of now, the technique for effectively embedding time series data that encompasses both spatial and temporal representations before inputting it into LLMs is not well-defined. 

\subsection{Traffic Prediction}

Traffic prediction aims to predict future traffic features based on historical traffic data, which is a crucial component in intelligent transportation systems~\cite{DBLP:conf/aaai/YeSDF021,10.14778/3551793.3551827,DBLP:journals/tkde/MiaoSCXW23}.
Traffic data is a special type of time series data. Thus, it is natural to adapt the classic time series models, such as ARIMA and Kalman filter, for the traffic prediction tasks in the early stage~\cite{kumar2015short}. 
Kumar et al. used the seasonal ARIMA model for short-term traffic prediction~\cite{kumar2015short}.
Chang et al. proposed a tensor-extended Kalman filter framework to characterize nonlinear dynamics and applied it to traffic forecasting~\cite{chang2023tensor}.
However, these models do not perform well due to the inherent spatial-temporal dependencies of traffic data.
Later, numerous efforts have been dedicated to advancing traffic prediction techniques by developing various neural network-based models.
In the beginning, convolutional neural networks (CNNs) were applied to traffic data to capture spatial dependencies in the data~\cite{shen2018stepdeep}. 
Shen et al. divided the city into grids and applied 3D CNN for traffic prediction~\cite{yuan2018hetero}. 
Yuan et al. used a convolutional long short-term memory network for traffic prediction~\cite{yuan2018hetero}.
Since CNNs are primarily designed to be applied in regular, grid-like urban areas, they encounter challenges when dealing with the non-Euclidean spatial structure of traffic data. This irregularity makes it difficult for CNNs to accurately capture the spatial dependencies inherent in traffic data.

Thanks to the apace of graph learning, graph convolutional network (GCN) based models are popular due to their permutation-invariance, local connectivity, and compositionally~\cite{song2020spatial,DBLP:conf/ijcai/WuPLJZ19,bai2020adaptive}. 
Li et al. modeled the traffic data as a directed graph and introduced a diffusion convolutional recurrent network for traffic prediction~\cite{li2018dcrnn_traffic}.
Choi et al. presented a graph neural controlled differential equation for traffic prediction~\cite{choi2022graph}. GCN-based models suffer from over-smoothing, making it hard to capture global spatial dependencies~\cite{DBLP:conf/aaai/JiangHZW23}.
More recently, attention-based models have emerged as a dominant trend~\cite{DBLP:conf/aaai/GuoLFSW19,zheng2020gman,9346058,DBLP:conf/aaai/LinLZCY20,xue2022translating,DBLP:conf/aaai/JiangHZW23}. Without taking the adjacency matrix into account, attention-based models can still model dynamic spatial correlation more effectively than GCN-based models. 
In~\cite{9346058}, the authors developed an attention-based spatial-temporal graph neural network for traffic prediction. However, the structures of these models are becoming increasingly sophisticated.

\begin{figure*}[htbp]
    \begin{centering}		
        \includegraphics[width=\textwidth]{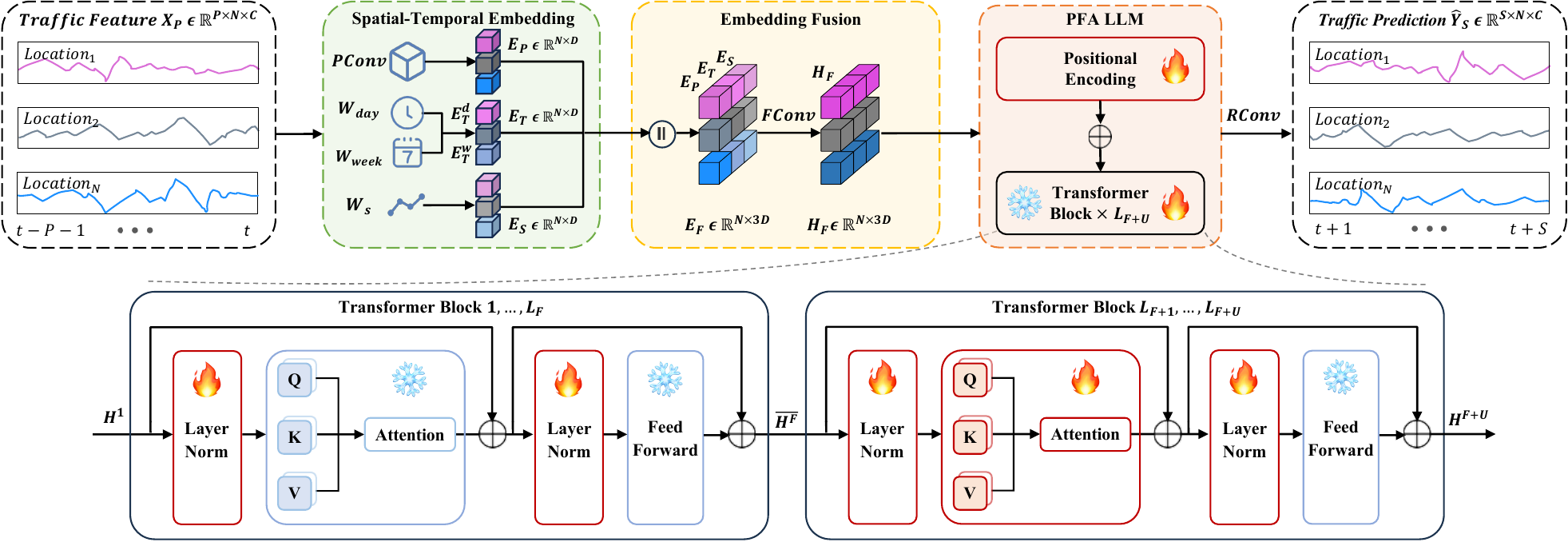}
        \caption{ST-LLM framework. Given an input traffic feature, we first embed it via a \textbf{Spatial-Temporal Embedding}. These embeddings are then integrated uniformly by an \textbf{Embedding Fusion} layer. The \textbf{PFA (partially frozen attention) LLM} has $F + U$ layers, which are divided into the first $F$ layers and the last $U$ layers. The multi-head attention and feed-forward layers in the first $F$ layers are frozen, and the multi-head attention in the last $U$ layers are unfrozen. The output from PFA LLM is regressed to the prediction results.
        }
        \label{fig: Framework}
    \end{centering}
\end{figure*}	
 
\section{Problem Definition}
\label{sec: definition}

\textit{Definition~1 (Traffic Feature).} We denote the traffic data as a tensor $\mat{X} \in \mathbb{R}^{T \times N \times C}$, where $T$ is the number of timesteps, $N$ is the number of spatial stations, and $C$ is the feature. For example, $C=1$ represents the traffic pick-up or drop-off flow.

\textit{Definition~2 (Traffic Prediction).} Given the historical traffic feature of $P$ timesteps $\mat{X}_P = \{\mat{X}_{t-P+1}, \mat{X}_{t-P+2}, \dots, \mat{X}_{t}\} \in \mathbb{R}^{P \times N \times C}$, the objective is to learn a function $f(\cdot)$ with parameter $\theta$ to predict traffic feature of on the following $S$ timesteps $\mat{Y}_S = \{\mat{Y}_{t+1}, \mat{Y}_{t+2}, \dots, \mat{Y}_{t+S}\} \in \mathbb{R}^{S \times N \times C}$. That is,
\begin{equation}
[\mat{X}_{t-P+1}, \mat{X}_{t-P+2}, \dots, \mat{X}_{t}] \xrightarrow[\theta]{f(\cdot)} [\mat{Y}_{t+1}, \mat{Y}_{t+2}, \dots, \mat{Y}_{t+S}],
\end{equation}

\noindent where each $\mat{X}_{i} \in \mathbb{R}^{ N \times C}$.

\section{Methodology}
\label{sec: methodology}

In this section, we provide a detailed elaboration of the proposed ST-LLM and its components. 

\subsection{Overview}

The Spatial-Temporal Large Language Model (ST-LLM) framework, as depicted in Figure~\ref{fig: Framework}, integrates a spatial-temporal embedding layer, a fusion convolution layer, an LLM layer, and a regression convolution layer. Initially, the historical traffic data is denoted as $\mat{X}_P$, which contains $N$ tokens of spatial locations. The $\mat{X}_P$ is processed through the spatial-temporal embedding layer, which extracts the token embedding of historical $P$ timesteps, spatial embedding, and temporal embedding, as $\mat{E}_T \in \mathbb{R}^{N \times D}$, $\mat{E}_S \in \mathbb{R}^{ N \times D}$, and $\mat{E}_P \in \mathbb{R}^{N \times D}$, respectively. A fusion convolution then integrates these representations into a unified way $\mat{E}_F \in \mathbb{R}^{N \times 3D}$. Subsequently, the $\mat{E}_F$ is input into a PFA LLM that encompasses $L+U$ layers, where the multi-head attention and feed forward layers in the first $F$ layers are frozen to preserve the pretrained knowledge and the multi-head attention layers in the last $U$ layers are unfrozen to enhance the model's focus on capturing the spatial-temporal dependencies between tokens, resulting in the output $\mat H^L \in \mathbb{R}^{N \times 3D}$. Finally, the regression convolution layer takes $\mat H^L$ and predicts the following traffic data, denoted as $\widehat{\mat{Y}}_S \in \mathbb{R}^{S \times N \times C}$.

\subsection{Spatial-Temporal Embedding and Fusion}

We aim to modify LLMs already trained for traffic prediction tasks. We define the timesteps at each location of traffic data as tokens. The spatial-temporal embedding layer transforms the tokens into spatial-temporal representations that align with the LLMs. These representations include spatial correlations, hour-of-day, day-of-week patterns, and token information.

We embed the tokens through a pointwise convolution, where the input data $\mat{X}_P$ is transformed into the embedding $\mat{E}_P \in \mathbb{R}^{N \times D}$:
\begin{equation}
    \mat{E}_P = \mathit{PConv}(\mat{X}_P; \theta_p),
    \label{eq: token emb}
\end{equation}

\noindent where $\mat{E}_P$ represents the token embedding. $\mathit{PConv}$ denotes the pointwise convolution operation using filters with a $1 \times 1$ kernel size. $\mat{X}_P$ is the input data, $D$ is the hidden dimension. $\theta_p$ represents the learnable parameters of the pointwise convolution.

To preserve the temporal information in the tokens, we utilize a linear layer to encode the input data into separate embeddings for the hour-of-day and day-of-week temporal embeddings. We perform absolute positional encoding for each traffic data at the ``day'' and ``week'' resolutions, and the generated positional encodings are $\mat X_{day} \in \mathbb{R}^{N \times T_d}$ and $\mat X_{week} \in \mathbb{R}^{N \times T_w}$. The hour-of-day embedding $\mat{E}_T^{d} \in \mathbb{R}^{N \times D }$ and day-of-week embedding $\mat{E}_T^{w}\in \mathbb{R}^{N \times D }$ are calculated as follows:
\begin{equation}
    \mat{E}_T^{d} = \mat W_{day} (\mat X_{day}),
\label{eq: te_1}
\end{equation}
\begin{equation}
    \mat{E}_T^{w} = \mat W_{week} (\mat X_{week}),
\label{eq: te_2}
\end{equation}
\begin{equation}
    \mat{E}_T = \mat{E}_T^{d} + \mat{E}_T^{w},
\label{eq: e_t}
\end{equation}
\noindent where $\mat{W}_{day} \in \mathbb{R}^{T_d \times D}$ and $\mat {W}_{week} \in \mathbb{R}^{T_w \times D}$ are the learnable parameter embeddings for the hour-of-day and day-of-week, respectively. By adding these two embeddings, we obtain the temporal representation $\mat{E}_T \in \mathbb{R}^{N \times D}$.

To represent spatial correlations among token pairs, we design an adaptive embedding of tokens, $\mat{E}_S \in \mathbb{R}^{ N \times D}$:
\begin{equation}
    \mat{E}_S = \sigma(\mat{W}_s \cdot \mat{X_P} + \mat b_s),
    \label{eq: se}
\end{equation}
 
\noindent where $\sigma$ denotes the activation function, $\mat{W}_s \in \mathbb{R}^{D \times D}$ and $ \mat b_s \in \mathbb{R}^D$ are the learnable parameter. 

Subsequently, we introduce a fusion convolution (FConv) to project the traffic feature to the required dimensions of the LLM. Specifically, the FConv integrates the token, spatial, and temporal embeddings to represent each token uniformly:
\begin{equation}
    \mat{H}_F = \mathit{FConv}(\mat{E}_P || \mat{E}_S || \mat{E}_T; \theta_f),
    \label{eq: fusion}
\end{equation}

\noindent where $\mat{H}_F \in \mathbb{R}^{N \times 3D}$, $||$ denotes concatenation, and $\theta_f$ represents the learnable parameters of the FConv.
 
\subsection{Partially Frozen Attention (PFA) LLM}

The frozen pretrained transformer (FPT) has demonstrated effectiveness in various downstream tasks across non-language modalities~\cite{DBLP:conf/aaai/LuGAM22}. However, its performance is less optimal in tasks requiring short-term and long-term predictions, such as traffic prediction~\cite{zhou2023onefitsall}. In this study, we propose a partially frozen attention (PFA) LLM, specifically designed to enhance prediction accuracy in traffic prediction.

The difference between the FPT and our PFA primarily lies in the frozen attention layers. In the FPT framework, both the multi-head attention and feed-forward layers are frozen during training, as these layers contain the most significant portion of the learned knowledge within the LLM.
In the PFA, we maintain the first $F$ layers identical to the FPT, but crucially, we unfreeze the last $U$ multi-head attention layers since the attentions effectively handle spatial-temporal dependencies in data. Consequently, our PFA LLM can adapt to traffic prediction while preserving the foundational knowledge acquired during pretraining.

Furthermore, our PFA LLM inverts the traditional calculation dimension from temporal to spatial. This inversion is intentional and aligns with the operation of the partially frozen layers. By focusing on spatial dimensions, our model captures global dependencies more effectively than if we were to concentrate solely on temporal aspects. This shift is particularly relevant in traffic prediction, where spatial dynamics play a critical role in determining flow patterns. 

The PFA LLM is built using a Transformer-based architecture, and we choose GPT2~\cite{radford2019language}. The GPT2 largely follows the details of the OpenAI GPT model~\cite{radford2018improving} with some modifications. Notably, the layer normalization is positioned at the input of each sub-block, akin to a pre-activation in a residual network. Additionally, an additional layer normalization is added after the final multi-head attention. We visualize these two modifications in the lower part of Figure 1. Furthermore, we introduce a PFA strategy to adapt the GPT2 to capture the spatial-temporal dependencies of the fused tensor $\mat H_F$. 

In the first $F$ layers of the PFA LLM, we freeze the multi-head attention and feed-forward layers:
\begin{align}
& \mat{\bar{H}}^{i} =\mathit{MHA}\left(\mathit{LN}\left(\mat{H}^{i}\right)\right) + \mat{H}^{i}, \notag \\
& \mat{H}^{i+1}  =\mathit{FFN}\left(\mathit{LN}\left(\mat{\bar{H}}^{i}\right)\right) + \mat{\bar{H}}^{i}, \notag \\
\label{eq: fpt}
\end{align}

\noindent where the range of $i$ is from $1$ to $F-1$, and $\mat{H}^1 =\left[\mat{H}_F + \mat{PE}\right]$. $\mat{PE}$ represent the learnable positional encoding. $\mat{\bar{H}}^i$ represents the intermediate representation of the $i_{th}$ layer after applying the frozen multi-head attention (MHA) and the first unfrozen layer normalization (LN). $\mat{H}^i$ symbolizes the final representation after applying the unfrozen LN and frozen feed-forward network (FFN).

The LN, MHA, and FFN in the PFA LLM are defined as follows:
\begin{align}
& \mathit{LN}\left(\mat{H}^{i}\right) = \gamma \odot \frac{\mat{H}^{i} - \mu}{\sigma} + \beta,  \notag \\
& \mathit{MHA}(\mat{\tilde H}^i) = \mat W^O (\mat{head}_1 || \cdots ||\mat{head}_h),  \notag \\
& \mat{head}_i = \mathit{Attention}(\mat W_i^Q \mat{\tilde H}^i, \mat W_i^K \mat{\tilde H}^i, \mat W_i^V \mat{\tilde H}^i), \notag \\
& \mathit{Attention}(\mat{\tilde H}^i)=\operatorname{softmax}\left(\frac{\mat{\tilde H}^i \mat{\tilde H}^{iT}}{\sqrt{d_k}}\right) \mat{\tilde H}^i, \notag \\ 
& \mathit{FFN}(\mat{\hat H}^i) = \max(0, \mat W_1\mat{\hat{H}}^{i+1}_P + \mat b_1)\mat W_2 + \mat b_2, \notag \\
\label{eq: gpt2}
\end{align}

\noindent where $\mat{\tilde H}^i$ is the output of $\mat{H}^{i}$ after passing through the first LN. $\mat{\hat H}^i$ is the output of $\mat{\bar{H}}^{i}$ after the second LN. $\gamma$ and $\beta$ are learnable scaling and translation parameters. $\mu$ and $\sigma$ represent the mean and standard deviation, respectively. $\odot$ denotes element-wise multiplication.

In the last $U$ layers of the LLM, we unfreeze the $\mathit{MHA}$ to adapt the ST-LLM for capturing spatial-temporal dependencies of traffic data:
\begin{align}
& \mat{\bar{H}^{F+U-1}} =\mathit{MHA}\left(\mathit{LN}\left(\mat{H^{F+U-1}}\right)\right) + \mat{H^{F+U-1}}, \notag \\
& \mat{H^{F+U}} =\mathit{FFN}\left(\mathit{LN}\left(\mat{\bar{H}^{F+U-1}}\right)\right) + \mat{\bar{H}^{F+U-1}},
\label{eq: pfa}
\end{align}

\noindent where $\mat{\bar{H}}^{F+U}$ represents the intermediate representation of the $L_{F+U-1}$ layer after applying the unfrozen MHA and the second frozen LN. $\mat{H}^{F+U}$ denotes the final output of the $L_{F+U}$ layer after applying both the unfrozen LN and frozen FFN, with the MHA being unfrozen.

After the PFA LLM, we design a regression convolution (RConv) to predict the traffic features on the following $S$ timesteps:
\begin{equation}
    \hat{\mat{Y}}_S = \textit{RConv}(\mat{H}^{F+U}; \theta_r),
    \label{eq: rconv}
\end{equation}

\noindent where $\widehat{\mat{Y}}_S \in \mathbb{R}^{S \times N \times C}$, and $\theta_r$ represents the learnable parameters of the regression convolution.

The loss function of ST-LLM is established as follows:
\begin{equation}
\mathcal{L} = \left\|\mat{\widehat {Y}}_S-\mat{Y}_S \right\| + \lambda \cdot L{\text{reg}},
\label{eq: loss}
\end{equation}

\noindent where $\mat{\widehat {Y}}_S$ is the predicted traffic feature. $\mat{Y}_S$ is the ground truth. $L{\text{reg}}$ represents the L2 regularization term, which helps control overfitting. $\lambda$ is a hyperparameter. The whole process of the ST-LLM is shown in Algorithm 1.

\begin{algorithm}
\SetAlgoLined
\DontPrintSemicolon
\caption{The ST-LLM Framework}
\KwIn{Traffic feature $\mat X_P$ in the historical $P$ timesteps, and all hyperparameters.}
\KwOut{Trained ST-LLM.}
\For{each epoch}{
    Shuffle training data\;
    \For{each batch $\mat X_P$ in training data}{
        $\mat E_F \leftarrow$ Spatial-Temporal Embedding by Equations (\ref{eq: token emb}), (\ref{eq: te_1}), (\ref{eq: te_2}) and (\ref{eq: se}) with $\mat X_P$.\;
        $\mat H_F~\leftarrow$ Embedding Fusion by Equation (\ref{eq: fusion}) with $\mat E_F$.\;
        \For{$i = 1$ \KwTo $F + U$}{
            $\mat H^1~\leftarrow$ PFA LLM Initialization with $\mat H_F$.\;
            \uIf{$i \leq F$}{
                calculate $\mat H^{i+1}$ by Equation~(\ref{eq: fpt}) with $\mat H^i$.
            }
            \Else{
                calculate $\mat H^{F+U}$ by Equation~(\ref{eq: pfa}) with $\mat H^{i}$.
            }
        }
        $\widehat{\mat Y}_S~\leftarrow$ by Equation~(\ref{eq: rconv}).\;
        Update all learnable parameters by minimizing the loss in Equation~(\ref{eq: loss}) with $\mat{\widehat {Y}}_S$ and $\mat{Y}_S$ via Ranger21 optimizer.\;
    }
}
\label{alg1}
\end{algorithm}

\section{Experiments}
\label{sec: experiments}

In this section, we aim to validate the superiority of our ST-LLM through a series of extensive experimental evaluations.

\subsection{Datasets}

This section details the datasets employed to examine the predictive performance of the ST-LLM and baselines, with real-world traffic data from NYCtaxi\footnote{\url{https://www.nyc.gov/site/tlc/about/tlc-trip-record-data.page}}, CHBike\footnote{\url{https://citibikenyc.com/system-data}}.

\begin{table}[htbp]
\caption{Dataset Description.}
\begin{tabular}{lcc}
\toprule
\textbf{Dataset Description}             & \textbf{NYCTaxi}      & \textbf{CHBike}       \\ \midrule
\textbf{Total Trips}         & 35 million            & 2.6 million           \\
\textbf{Number of Stations}  & 266                   & 250                   \\
\textbf{Time Span}           & 01/04/2016~-~30/06/2016 & 01/04/2016~-~30/06/2016 \\
\textbf{Number of Timesteps} & 4,368                 & 4,368                 \\
\textbf{Timestep Interval}   & 30 minutes            & 30 minutes            \\ \bottomrule
\end{tabular}
\end{table}

\textbf{NYCTaxi}. The NYCTaxi dataset comprises over 35 million taxi trips in New York City (NYC), systematically categorized into 266 virtual stations. Spanning three months from April 1st to June 30th, 2016, it includes 4,368 timesteps, each representing a half-hour interval.

\textbf{CHBike}. Consisting of approximately 2.6 million Citi bike orders, the CHBike dataset reflects the usage of the bike-sharing system in the same period as the NYCTaxi dataset, from April 1st to June 30th, 2016. After filtering out stations with few orders, it focuses on the 250 most frequented stations. The dataset aligns with NYCTaxi in terms of time, covering 4,368 timesteps with each timestep representing a 30-minute interval.

\subsection{Baselines}
We compare ST-LLM with the following 10 baselines belonging to three categories: (1) GNN-based models: DCRNN~\cite{li2018dcrnn_traffic}, STGCN~\cite{10.5555/3304222.3304273}, GWN~\cite{DBLP:conf/ijcai/WuPLJZ19}, AGCRN~\cite{bai2020adaptive}, STG-NCDE~\cite{choi2022graph}, DGCRN~\cite{DBLP:journals/tkdd/LiFYJYSJL23}.
(2) Attention-based models: ASTGCN~\cite{DBLP:conf/aaai/GuoLFSW19}, 
GMAN~\cite{zheng2020gman}, ASTGNN~\cite{9346058}. (3) LLMs: OFA~\cite{zhou2023onefitsall}, GATGPT~\cite{chen2023gatgpt}, GCNGPT, and LLAMA2. The details of the baselines are outlined as follows:

\begin{itemize}
    \item DCRNN~\cite{li2018dcrnn_traffic}: it models the data as a directed graph, and introduces diffusion convolutional recurrent network.
    \item STGCN~\cite{10.5555/3304222.3304273}: A graph convolutional network that combines 1D convolution to tackle the time series prediction task in the traffic domain.
    \item GWN~\cite{DBLP:conf/ijcai/WuPLJZ19}: A graph neural network that employs graph convolution with an adaptive adjacency matrix.
    \item AGCRN~\cite{bai2020adaptive}: it is an adaptive graph convolutional recurrent network that incorporates node learning and inter-dependency inference among traffic series. 
    \item STG-NCDE~\cite{choi2022graph}: it presents the graph neural controlled differential equation for processing sequential data.
    \item DGCRN~\cite{DBLP:journals/tkdd/LiFYJYSJL23}: it introduces a traffic prediction framework using dynamic graph convolutional recurrent networks.
    \item ASTGCN~\cite{DBLP:conf/aaai/GuoLFSW19}: it is an attention-based spatial-temporal graph convolutional network for traffic forecasting.
    \item GMAN~\cite{zheng2020gman}: it is an attention-based predictive model that adopts an encoder-decoder architecture.
    \item ASTGNN~\cite{9346058}: it is an attention-based model for learning the dynamics and heterogeneity of traffic data.
    \item OFA~\cite{zhou2023onefitsall}: it refrains from altering the self-attention and feed-forward networks of the residual blocks in the GPT2. We take an inverted view on traffic data of OFA for better prediction performance.
    \item GATGPT~\cite{chen2023gatgpt}: it combines the GAT with the frozen pretrained transformer GPT2.
    \item GCNGPT: it combines the GCN with the frozen pretrained transformer GPT2.
    \item LLAMA2: it is a collection of pretrained and fine-tuned large language models developed by Meta. In the LLAMA2, we adapt the frozen pretrained transformer.
\end{itemize}

\subsection{Implementations}

Aligning with contemporary practices, we divided the NYCTaxi and CHBike datasets into training, validation, and test sets using a 6:2:2 ratio. We set the historical timesteps $P$ and the future timesteps $S$ to 12 each, enabling multi-step traffic prediction. $T_w$ is set at 7 to represent a week's seven days. $T_d$ is 48, with each timestep spanning 30 minutes. The experiments were carried out on a system incorporating NVIDIA A100 GPUs, each with 40GB of memory. For training LLM-based models, we used the Ranger21 optimizer with a learning rate of 0.001, while GCN and attention-based models employed the Adam optimizer, also set at a 0.001 learning rate. The LLMs used are GPT2 and LLAMA2 7B. We configured GPT2 with six layers~\cite{zhou2023onefitsall}, and LLAMA2 with eight layers~\cite{jin2023time}. The experiments were conducted with a batch size of 64 over 100 training epochs. Note that the experimental results are averaged across all prediction timesteps. 

\subsection{Evaluation Metrics}

Four metrics were used for evaluating the models: Mean Absolute Error (MAE), Mean Absolute Percentage Error (MAPE), Root Mean Squared Error (RMSE), and Weighted Absolute Percentage Error (WAPE). MAE and RMSE quantify absolute errors, while MAPE and WAPE assess relative errors. In all metrics, lower values indicate superior prediction performance:
\begin{small}
\begin{equation}
\text{MAE} = \frac{1}{m}\sum_{i=1}^{m} \left| \mat{\widehat {Y}}_{i} - \mat{Y}_{i} \right|,~~
\text{MAPE} = \frac{100\%}{m}\sum_{i=1}^{m} \left| \frac{\mat{\widehat {Y}}_{i} - \mat{Y}_{i}}{\mat{Y}_{i}} \right|,
\end{equation}
\begin{equation}
\text{RMSE} = \sqrt{\frac{1}{m}\sum_{i=1}^{m} \left( \mat{\widehat {Y}}_{i} - \mat{Y}_{i} \right)^2},~
\text{WAPE} = \frac{\sum_{i=1}^{m} \left| \mat{\widehat {Y}}_{i} - \mat{Y}_{i} \right|}{\sum_{i=1}^{m} \left| \mat{Y}_{i} \right|} \times 100\%,
\end{equation}
\end{small}

\noindent where $m$ is the number of all predicted values.

\subsection{Main Results}    

\begin{table*}[htbp]
\caption{Model comparison on NYC datasets in terms of MAE, RMSE, MAPE (\%), and WAPE (\%). Results are averaged from all prediction timesteps.}
\label{tab: comparison on nyc}
\resizebox{\textwidth}{!}{
\begin{tabular}{c|cccc|cccc|cccc|cccc}
\hline
Dataset   & \multicolumn{4}{c|}{NYCTaxi Pick-up}                             & \multicolumn{4}{c|}{NYCTaxi Drop-off}                               & \multicolumn{4}{c|}{CHBike Pick-up}                                & \multicolumn{4}{c}{CHBike Drop-off}                             \\ \hline
Metric   & MAE           & RMSE       & MAPE             & WAPE             & MAE           & RMSE          & MAPE             & WAPE             & MAE           & RMSE          & MAPE             & WAPE             & MAE           & RMSE          & MAPE          & WAPE             \\ \hline
DCRNN    & 5.40          & 9.71       & 35.09\%          & 20.43\%          & 5.19          & 9.63          & 37.78\%          & 19.82\%          & 2.09          & 3.30          & 54.22\%          & 42.26\%          & 1.96          & 2.94          & 51.42\%       & 39.61\%          \\
STGCN    & 5.71          & 10.22      & 36.51\%          & 21.62\%          & 5.38          & 9.60          & 39.12\%          & 20.55\%          & 2.08          & 3.31          & 53.63\%          & 42.08\%          & 2.01          & 3.07          & 50.45\%       & 40.62\%          \\
ASTGCN   & 7.43          & 13.84      & 47.96\%          & 28.04\%          & 6.98          & 14.70         & 45.48\%          & 26.60\%          & 2.76          & 4.45          & 64.23\%          & 55.71\%          & 2.79          & 4.20          & 69.88\%       & 56.49\%          \\
GWN      & 5.43          & \textbf{9.39}       & 37.79\%          & 20.55\%          & \textbf{5.03}    &  \textbf{8.78}    & 35.63\%          & {\ul19.21\%}    & {\ul 2.04}          & {\ul 3.20}          & \textbf{53.08\%}    & {\ul 40.95\%}    & {\ul 1.95}          & 2.98          & 50.30\% & {\ul 39.43\%}          \\
AGCRN    & 5.79          & 10.11      & 40.40\%          & 21.93\%          & 5.45          & 9.56          & 40.67\%          & 20.81\%          & 2.16          & 3.46          & 56.35\%          & 43.69\%          & 2.06          & 3.19          & 51.91\%       & 41.78\%          \\
GMAN     & 5.43          &  9.47       & {\ul 34.39\%}    & 20.42\%    & 5.09          & {\ul8.95}          & {\ul 35.00\%}    & 19.33\%          & 2.20          & 3.35          & 57.34\%          & 44.06\%          & 2.09          & 3.00          & 54.82\%       & 42.00\%          \\
STSGCN   & 6.19          & 11.14      & 39.67\%          & 25.37\%          & 5.62          & 10.21         & 37.92\%          & 22.59\%          & 2.36          & 3.73          & 58.17\%          & 50.09\%          & 2.73          & 4.50          & 57.89\%       & 54.10\%          \\
ASTGNN   & 5.90          & 10.71      & 40.15\%          & 22.32\%          & 6.28          & 12.00         & 49.78\%          & 23.97\%          & 2.37          & 3.67          & 60.08\%          & 47.81\%          & 2.24          & 3.35          & 57.21\%       & 45.27\%          \\
STG-NCDE & 6.24          & 11.25      & 43.20\%          & 23.46\%          & 5.38          & 9.74          & 40.45\%          & 21.37\%          & 2.15          & 3.97          & 55.49\%          & 61.38\%          & 2.28          & 3.42          & 60.96\%       & 46.06\%          \\
DGCRN    & 5.44          & 9.82       & 35.78\%          & 20.58\%          & 5.14          & 9.39          & 35.09\%          & 19.64\%          & 2.06          & 3.21          & 54.06\%          & 41.51\%          & 1.96          & {\ul 2.93}          & 51.99\%       & 39.70\%          \\
OFA   & 5.82          & 10.42      & 36.67\%          & 22.00\%          & 5.60          & 10.14         & 37.39\%          & 21.36\%          & 2.06          & 3.21          & 53.55\%          & 41.70\%          & 1.96          & 2.97          & {\ul 49.64\%}       & 39.68\%          \\
GATGPT & 5.92 & 10.55 & 37.83\% & 22.39\% & 5.66 & 10.39 & 37.36\% & 21.60\% & 2.07 & 3.23 & 52.54\% & 41.70\% & 1.95 & 2.94 & 49.26\% & 39.43\% \\
GCNGPT & 6.58 & 12.23 & 40.19\% & 24.88\% & 6.64 & 12.24 & 42.46\% & 25.32\% & 2.37 & 3.80 & 56.24\% & 47.66\% & 2.24 & 3.48 & 51.05\% & 45.37\%  \\
LLAMA2 & {\ul 5.35}   & 9.48   & 41.32\%  & {\ul 20.27\%}  & 5.66  & 10.74  & 47.47\%  & 21.63\%  & 2.10  & 3.37  & 56.63\%  & 42.49\%  & 1.99  & 3.03  & 55.23\%  & 40.28\% \\
ST-LLM   & \textbf{5.29} & {\ul 9.42}       & \textbf{33.55\%} & \textbf{20.03\%} & {\ul 5.07}          & 9.07          & \textbf{33.34\%} & \textbf{19.18}\%          & \textbf{1.99} & \textbf{3.08} & {\ul 53.54\%} & \textbf{40.19\%} & \textbf{1.89} & \textbf{2.81} & \textbf{49.50\%}       & \textbf{38.27\%} \\ \hline
\end{tabular}}
\end{table*}

The comparison results with baselines are shown in Table~\ref{tab: comparison on nyc}. The bold results are the best, and the underlined results are the second best. The LLM in ST-LLM is GPT2. 
We can make the following observations. (1) LLM-based methods yield superior prediction results, with ST-LLM exhibiting the most effective performance. The ST-LLM outperforms other LLMs in four traffic prediction scenarios, demonstrating its superior accuracy in handling diverse traffic data across various datasets. (2) OFA and LLAMA2 are competent but surpassed by ST-LLM which achieves a 22.5\% average MAE improvement over OFA and 20.8\% over LLAMA2. This may be due to OFA's ineffective traffic data embedding, making it difficult for LLMs to understand the spatial-temporal dependencies between data. Despite LLAMA2's larger size and complexity, it doesn't directly translate to better traffic prediction than ST-LLM. GATGPT and GCNGPT do not extract temporal representations of traffic data to influence the LLM to capture spatial-temporal dependencies. (3) Attention-based models like ASTGNN and GMAN exhibit varied performance across different datasets. They performed quite well in some cases but were always inferior to ST-LLM. This variability could be attributed to the limitations of traditional attention mechanisms in handling complex spatial-temporal embeddings, especially when compared to large language models. (4) GNN-based Models such as GWN and DGCRN demonstrate competitive performance, particularly in specific metrics, but still cannot outperform ST-LLM. This suggests that while GNNs effectively capture spatial dependencies, their temporal analysis capabilities might not be as advanced as the ST-LLM, which limits their overall performance.

In summary, the experimental results from the traffic prediction tasks using the NYCTaxi and CHBike datasets demonstrate a clear trend in performance among different types of models. LLMs-based methods emerge as the top performers, showcasing their superior ability to handle different traffic prediction tasks. Following LLM-based models, attention-based models occupy the second tier. Finally, GCN-based models, while still effective, rank lower compared to the aforementioned models. This hierarchy highlights the evolving landscape of model capabilities, with LLM-based methods leading the way in traffic prediction tasks.

\subsection{Performance of ST-LLM and Ablation Studies}

\textbf{With or Without Different Components.} The ST-LLM comprises several crucial components, each contributing to its overall effectiveness in traffic prediction. This section compares variants of ST-LLM concerning the following aspects to investigate the effectiveness of different components. w/o LLM: A variant of ST-LLM with the LLM being removed. w/o ST: A variant of ST-LLM with the spatial-temporal embedding being removed. w/o T: A variant of ST-LLM with the temporal embedding being removed. w/o S: A variant of ST-LLM with the spatial embedding being removed.

\begin{figure}[t]
    \centering
    \subfigure[Drop-off under MAPE and WAPE.]{
    \includegraphics[width=0.46\linewidth]{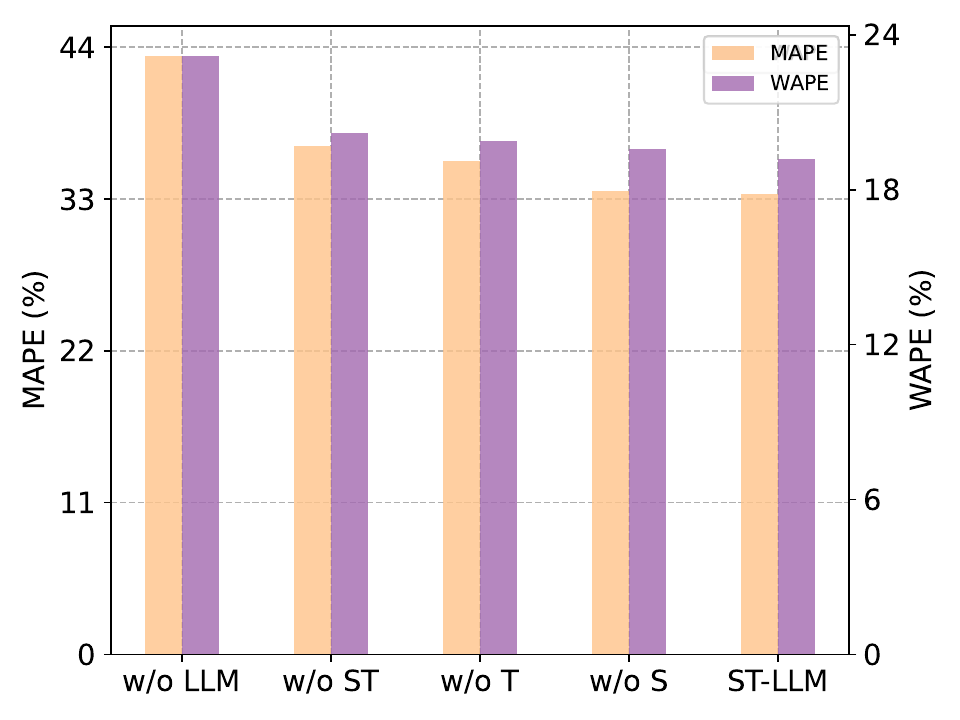}
    }\hfill
    \subfigure[Drop-off under MAE and RMSE.]{
    \includegraphics[width=0.46\linewidth]{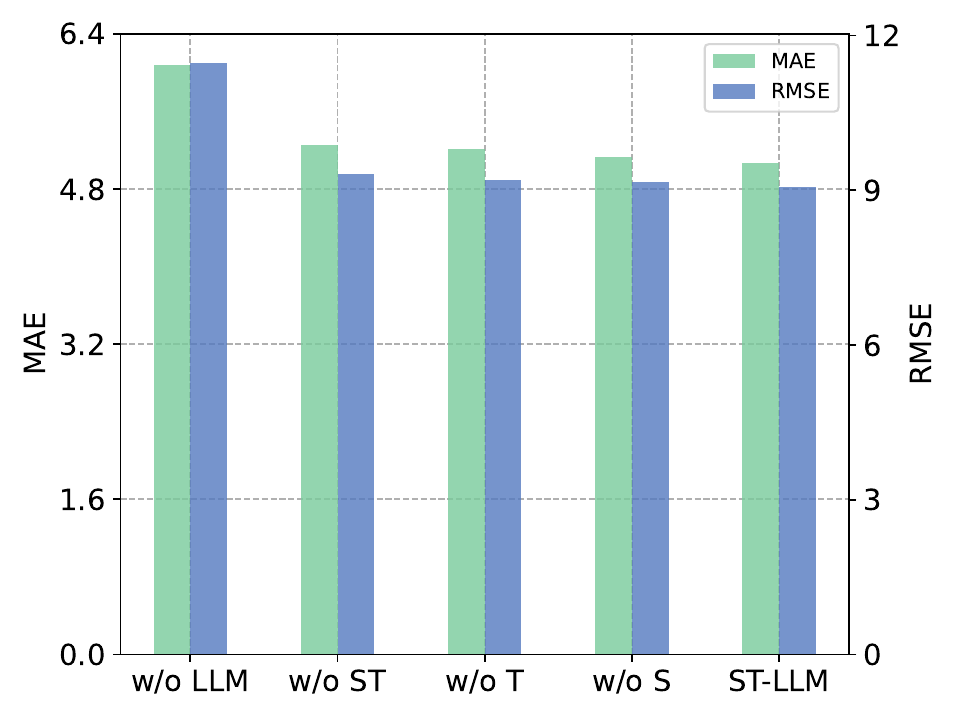}
    }\hfill
    
    \subfigure[Pick-up under MAPE and WAPE.]{
    \includegraphics[width=0.46\linewidth]{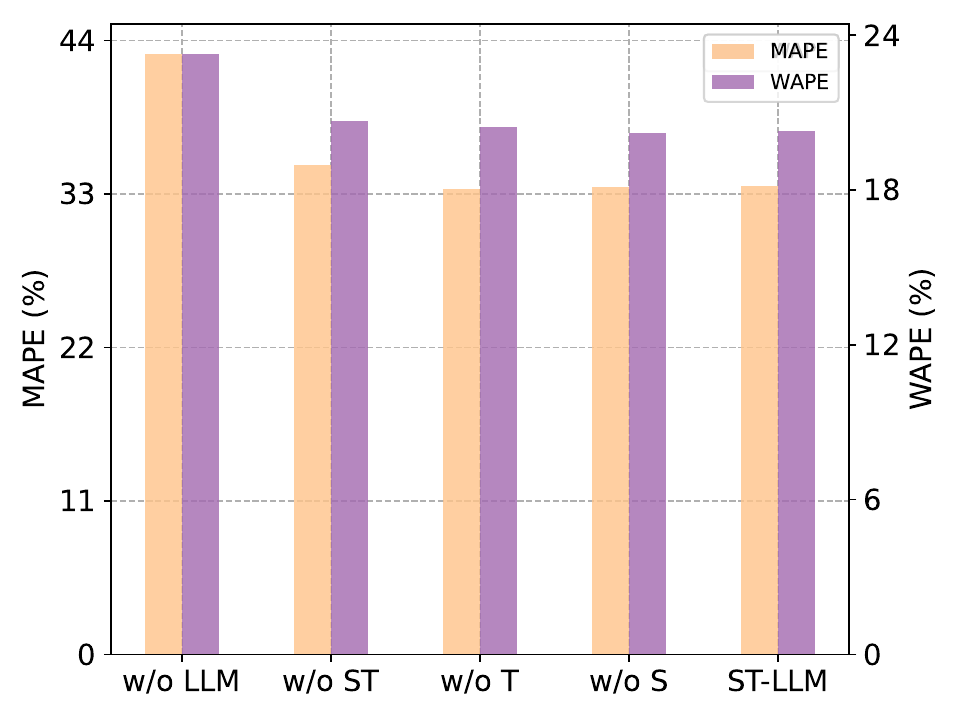}
    }\hfill
    \subfigure[Pick-up under MAE and RMSE.]{
    \includegraphics[width=0.46\linewidth]{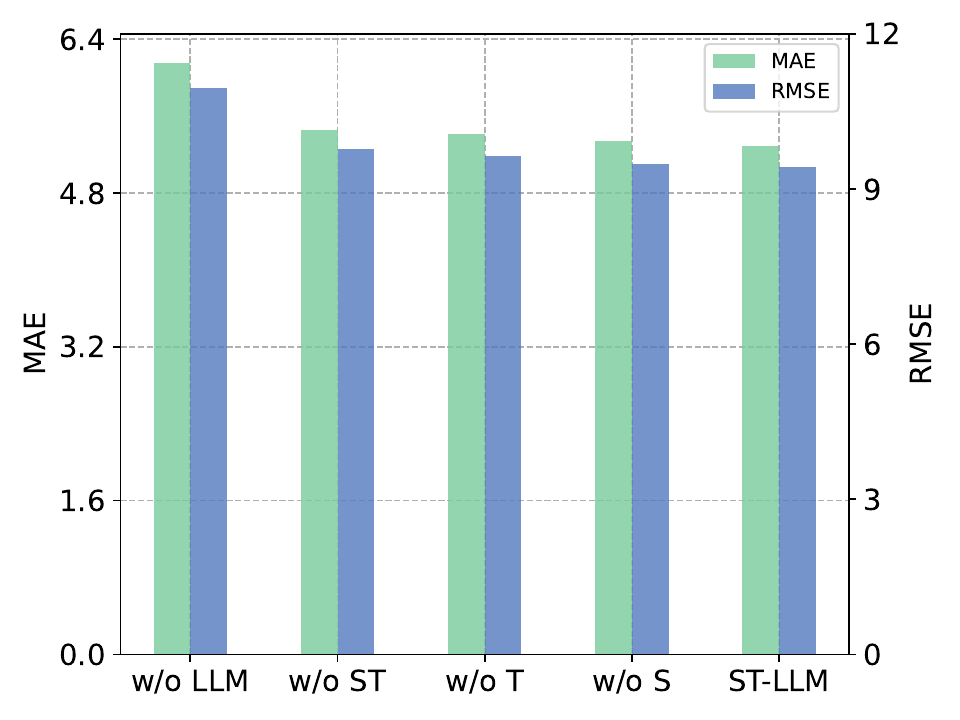}
    }\hfill
    \caption{Ablation study of ST-LLM on NYCTaxi dataset. 
    }
    \label{fig: albation}
\end{figure}

\begin{figure}[htbp]
    \centering
    \subfigure[NYCTaxi Pick-up under WAPE.]{
    \includegraphics[width=0.47\linewidth]{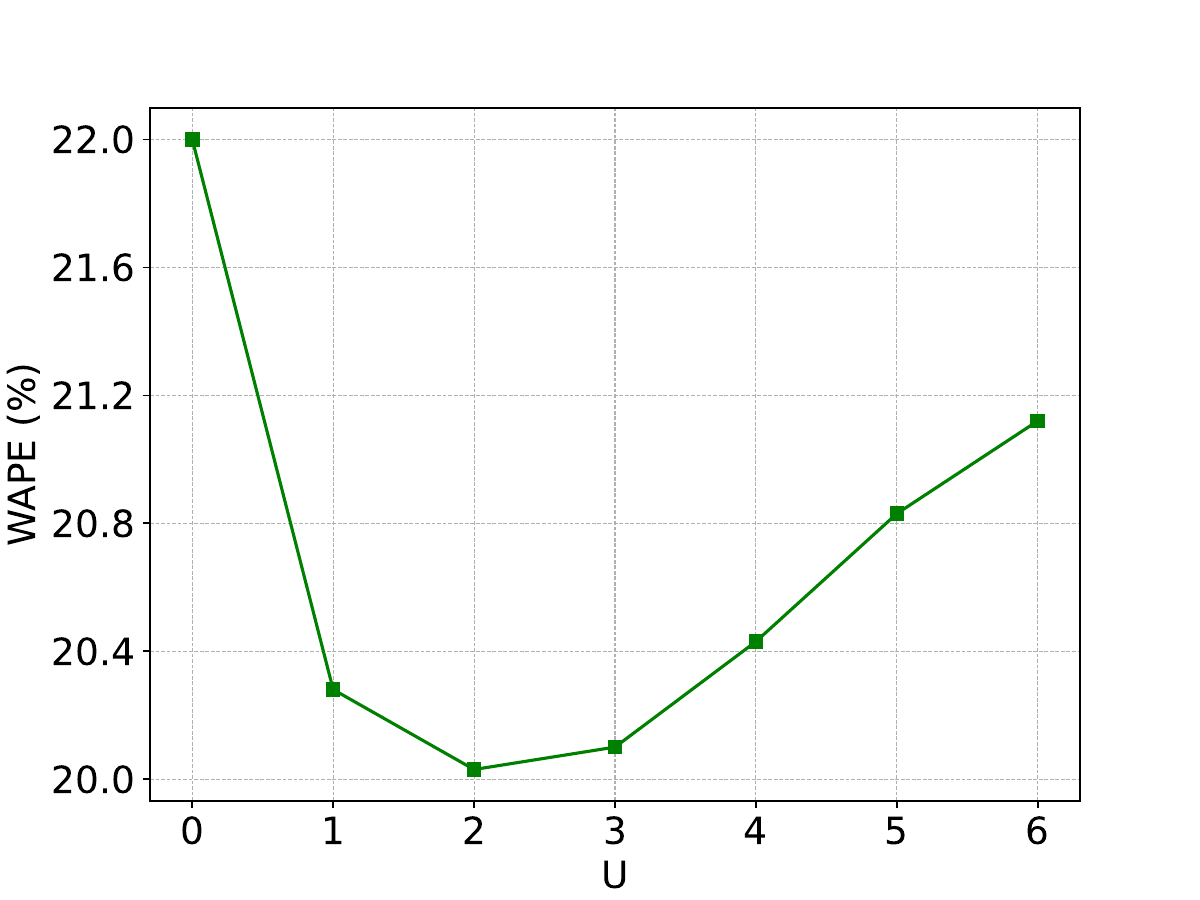}
    }\hfill
    \subfigure[NYCTaxi Pick-up under MAE.]{
    \includegraphics[width=0.47\linewidth]{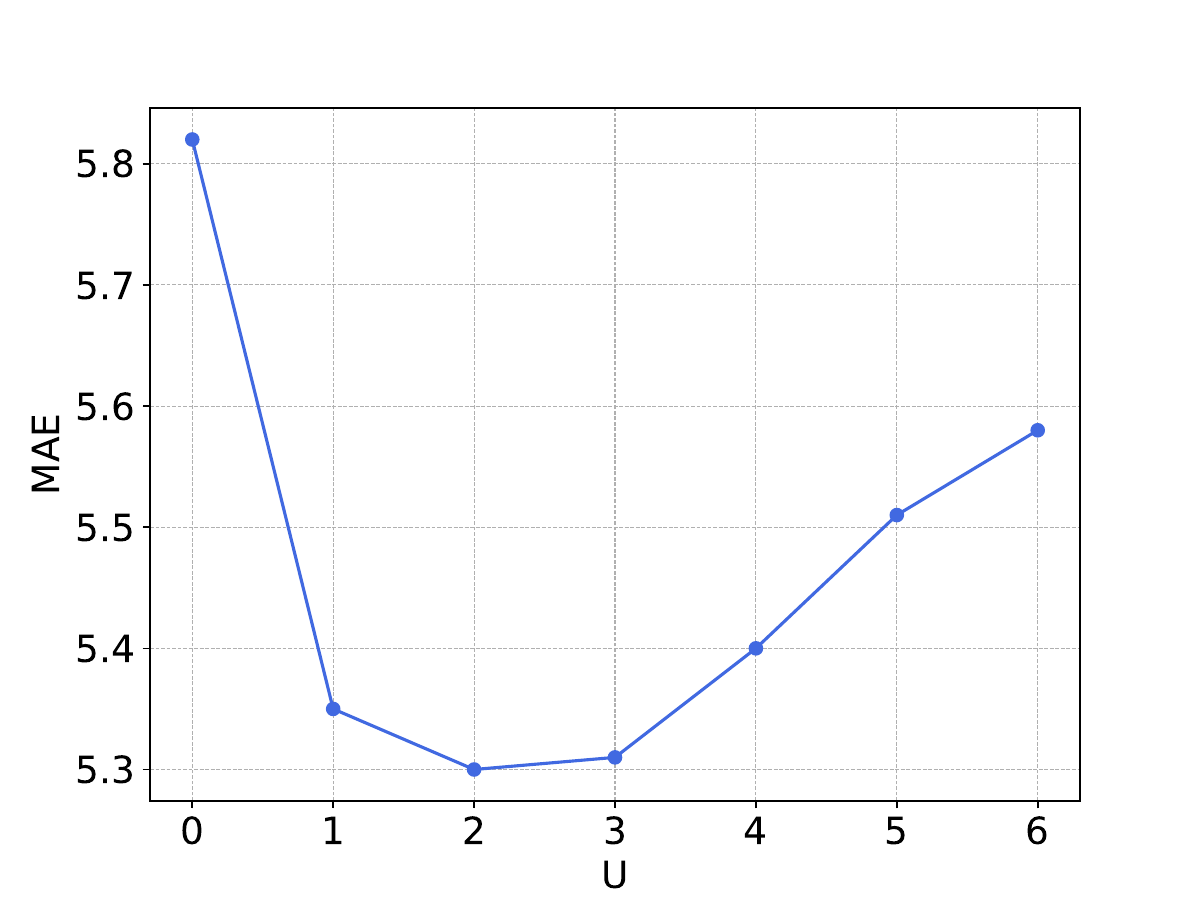}
    \hfill
    }
    
    \subfigure[CHBike Pick-up under WAPE.]{
    \includegraphics[width=0.47\linewidth]{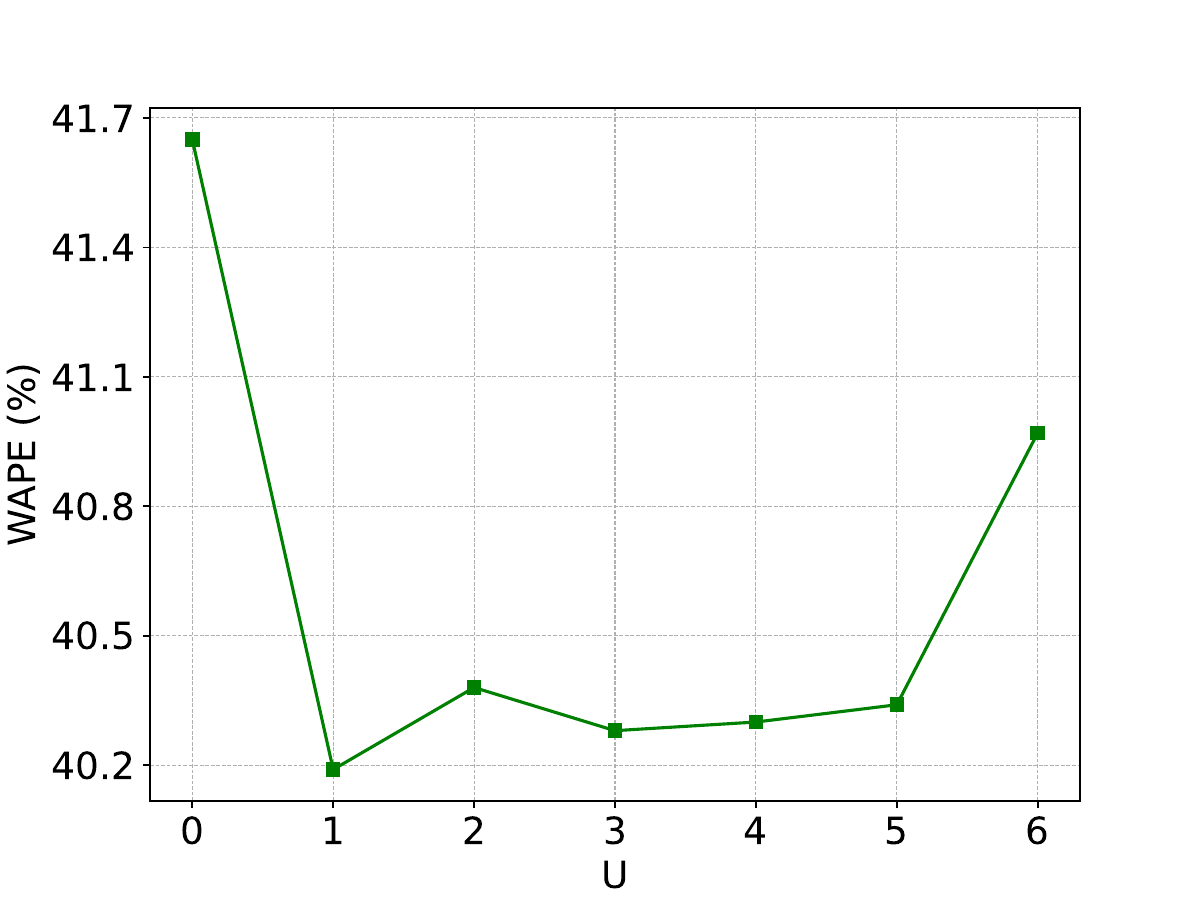}
    }\hfill
    \subfigure[CHBike Pick-up under MAE.]{
    \includegraphics[width=0.47\linewidth]{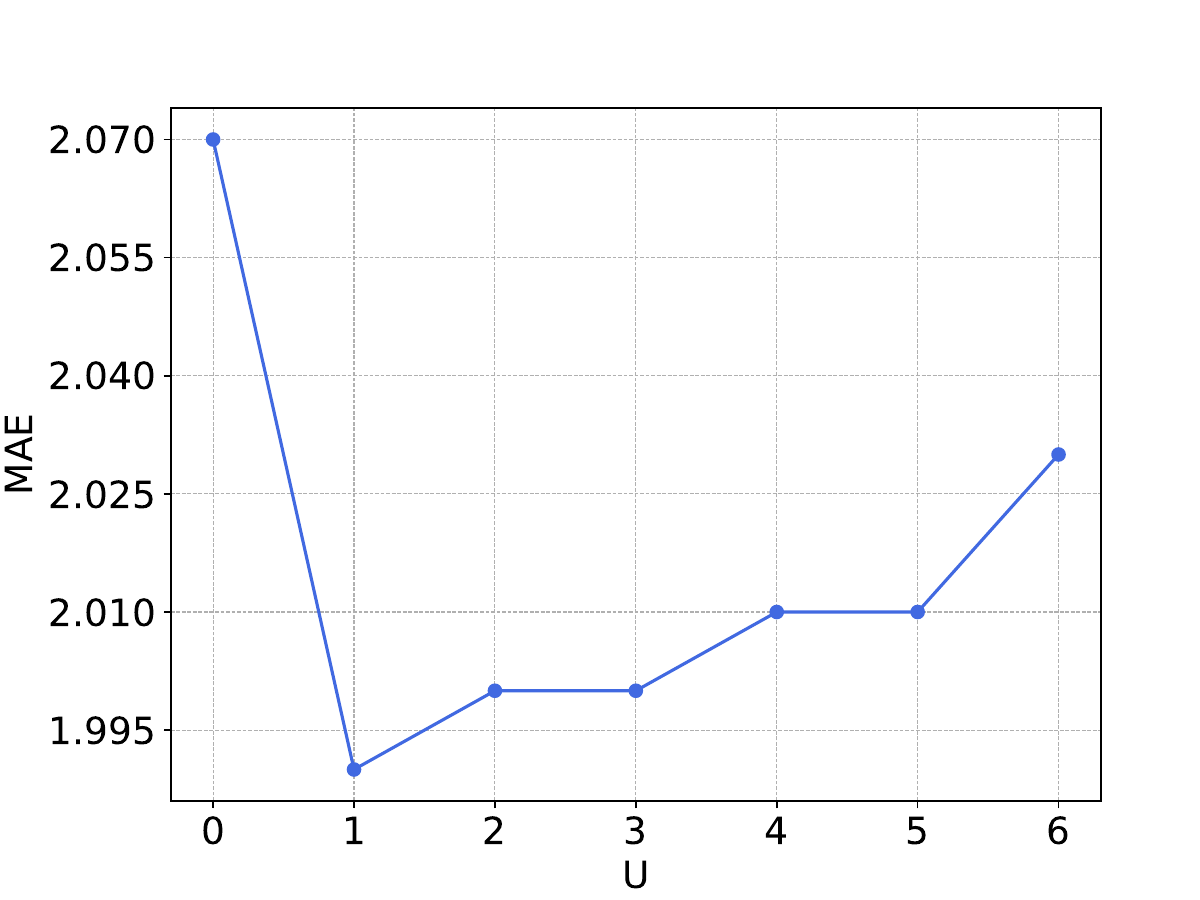}
    \hfill
    }
    \caption{Performance study of unfreezing last $U$ layers.}
    \label{fig: Parameter}
\end{figure}

\begin{table*}[htbp]
\small
\caption{Ablation Study of Partially Frozen Attention LLM.}
\resizebox{\textwidth}{!}{
\begin{tabular}{l|ccc|ccc|ccc|ccc|ccc}
\hline
\multicolumn{1}{c|}{LLM}    & \multicolumn{3}{c|}{No Pretrain} & \multicolumn{3}{c|}{Full Layer} & \multicolumn{3}{c|}{Full Tuning} & \multicolumn{3}{c|}{FPT} & \multicolumn{3}{c}{PFA}                          \\ \hline
\multicolumn{1}{c|}{Metric} & MAE      & RMSE     & WAPE       & MAE      & RMSE    & WAPE       & MAE      & RMSE     & WAPE       & MAE   & RMSE  & WAPE     & MAE           & RMSE          & WAPE             \\ \hline
NYCTaxi Drop-off             & 5.22     & 9.19     & 19.93\%    &  5.22     &   9.33  &  19.91\%    & 5.90     & 10.36     & 22.51\%    & 5.73  & 10.43  & 21.87\%  & \textbf{5.07} & \textbf{9.07} & \textbf{19.18\%} \\
NYCTaxi Pick-up              & 5.36     & 9.39     & 20.27\%    &  5.43    & 9.65    & 20.54\%    &  5.98     &  10.40     &  22.63\%    & 5.83  & 10.45  & 22.04\%  & \textbf{5.29} & \textbf{9.42} & \textbf{20.30\%} \\ 
CHBike Drop-off             & 1.92     & 2.86     & 38.84\%    & 1.91     & 2.83    & 38.63\%    & 1.90     & 2.82     & 38.28\%    & 1.92  & 2.86  & 38.90\%  & \textbf{1.89} & \textbf{2.81} & \textbf{38.27\%} \\
CHBike Pick-up              & 2.03     & 3.14     & 40.87\%    & 2.02     & 3.12    & 40.62\%    & 2.01     & 3.11     & 40.43\%    & 2.07  & 3.25  & 41.65\%  & \textbf{1.99} & \textbf{3.08} & \textbf{40.19\%} \\ 
\hline
\end{tabular}}
\label{tab: ablation of pfa}
\end{table*}

Figure~\ref{fig: albation} presents the ablation study on the NYCTaxi Pick-up and Drop-off datasets, examining the impact of different components in the ST-LLM. The w/o LLM variant shows a considerable increase in error across all metrics. Its removal leads to a degradation in performance, demonstrating that the prediction capabilities of the ST-LLM are heavily reliant on the LLM's ability to learn complex dependencies from traffic data.
The exclusion of spatial-temporal embedding (w/o ST) results in a notable performance drop. This highlights the importance of spatial-temporal embedding in understanding the spatial-temporal dependencies within the context of traffic data. The experimental results reveal that removing either temporal (w/o T) or spatial (w/o S) components similarly affects the model's prediction error. Ablating either of these embeddings results in an increased error, indicating the necessity of both for accurate predictions. Notably, the model incurs a larger prediction error without the temporal component, underscoring the significance of our thoughtfully designed hour-of-day and day-of-week embeddings. This observation further emphasizes the critical role that balanced spatial and temporal embeddings play in enhancing the model's predictive performance. We observe the lowest error rates across all metrics when all components are integrated, as in the full ST-LLM model. This underscores the effect of combining LLM, spatial, and temporal embeddings to handle the spatial-temporal dependencies of traffic prediction.

\textbf{Ablation Study of Partially Frozen Attention LLM.} In this subsection, we conducted an ablation study to evaluate the efficacy of our proposed Partially Frozen Attention (PFA) LLM. The PFA is compared against several variations: Frozen Pretrained Transformer (FPT), models without pretraining (No Pretrain), models utilizing the full twelve layers of GPT-2 (Full Layer), and fully tuned models without any frozen layers (Full Tuning). The ablation results of PFA LLM are shown in table~\ref{tab: ablation of pfa}. The PFA demonstrates superior performance across all metrics on all datasets. This suggests that partially freezing the attention significantly enhances the predictive accuracy. The FPT shows commendable performance, it is slightly outperformed by the PFA. This indicates that the partial freezing strategy strikes a more optimal balance between leveraging prelearned features and adapting to new data. The Full Layer and Full Tuning models exhibit competitive performance. However, they still fall short of the efficiency and accuracy demonstrated by the PFA model. This underscores the advantage of selective freezing in managing the adaptability of the model. The comparison with the No Pretrain model highlights the significant role of pretraining in model performance. While the No Pretrain model performs reasonably well, it is evident that pretraining, especially when combined with strategies like partial frozen, is crucial for achieving higher levels of accuracy.

\subsection{Parameter Analysis}

In the ST-LLM framework shown in Figure~\ref{fig: Parameter}, the hyperparameter $U$ plays a pivotal role in determining the count of unfrozen multi-head attention layers throughout the training phase. As depicted in Figure~\ref{fig: Parameter} (a), for the NYCTaxi Pick-up dataset, the performance, as measured by WAPE, initially improves with the increase of $U$ to 2. The trend suggests that unfreezing additional layers up to a certain threshold can enhance the performance of the ST-LLM. However, this positive effect inverts when $U$ exceeds 2, at which point the model's performance starts to degrade, hinting at the diminishing benefits associated with unfreezing more layers.
Figure~\ref{fig: Parameter} (b) presents a consistent pattern, where the MAE for the NYCTaxi Pick-up dataset decreases with an increase in $U$ to 2. In a nutshell, the optimal $U$ of taxi flow prediction is set to 2. 

For the CHBike Pick-up dataset, as shown in Figure ~\ref{fig: Parameter} (c), setting $U$ to 1 results in the lowest WAPE, signifying the peak performance of the model. An increase in $U$ leads to a rise in WAPE, which signals a decline in accuracy. Figure~\ref{fig: Parameter} (d) illustrates a similar pattern on the CHBike Pick-up dataset. The MAE indicates an increase in performance as $U$ is set to 1, with the lowest MAE observed at this value. This reinforces the observation that a single unfrozen multi-head attention layer is optimal for minimizing absolute errors, and the model achieves the balance between complexity and performance. This optimal point suggests that unfreezing more layers does not contribute to improved accuracy and might even degrade ST-LLM performance.

\begin{table*}[htbp]
\caption{Few-shot prediction results on 10\% data of LLMs.}
\label{tab: few shot}
\resizebox{\textwidth}{!}{
\begin{tabular}{c|cccc|cccc|cccc|cccc}
\hline
\multirow{2}{*}{LLM} & \multicolumn{4}{c|}{NYCTaxi Pick-up}                                                       & \multicolumn{4}{c|}{NYCTaxi Drop-off}                                                      & \multicolumn{4}{c|}{CHBike Pick-up}                                                        & \multicolumn{4}{c}{CHBike Drop-off}                                                                             \\ \cline{2-17} 
                     & MAE                  & RMSE                 & MAPE                 & WAPE                  & MAE                  & RMSE                 & MAPE                 & WAPE                  & MAE                  & RMSE                 & MAPE                 & WAPE                  & \multicolumn{1}{c}{MAE}  & \multicolumn{1}{c}{RMSE} & \multicolumn{1}{c}{MAPE}    & \multicolumn{1}{c}{WAPE}    \\ \hline
OFA                  &6.49 &12.12 &46.74\% &24.54\% &6.27 &12.10 &45.23\% &23.92\% & 2.20                 & 3.59                 & 57.52\%              & 44.40\% & 2.06 & 3.17 & 55.96\% & 41.63\%           \\

GATGPT & 7.02 & 13.09 & 50.19\% & 26.54\% & 6.84 & 13.27 & 56.15\% & 26.09\% & 2.59 & 4.41 &  56.23\% & 52.20\% & 2.50  & 4.07  &  56.36\% &  50.64\%           \\
GCNGPT                  & 10.31 & 18.82 & 59.41\% & 39.02\% & 9.25 & 19.50 & 56.77\% & 35.28\% & 2.73                  & 4.44                  &  56.93\%              &  55.20\% & 2.79 &  4.65 &  61.85\% & 56.28\%           \\
LLAMA2       & 5.81 & 10.16 & 41.82\% & 21.99\% & 5.59 & 9.90 & 40.58\% & 21.35\% & 2.24 & 3.58 & 59.47\% & 45.20\% &   2.11  & 3.23  &  54.44\% & 42.75\%                              \\
ST-LLM              & \textbf{5.40}                 & \textbf{9.63}                & \textbf{33.36\%}              & \textbf{20.45\%}               & \textbf{5.54}                 & \textbf{9.84}                 & \textbf{39.56\%}              & \textbf{21.14\%}               & \textbf{2.07}                 & \textbf{3.23}                 & \textbf{55.68\%}              & \textbf{41.85\%}               & \textbf{1.93}                     & \textbf{2.88}                     & \textbf{52.75\%}                     & \textbf{39.21\%}                     \\
\hline
\end{tabular}
}
\end{table*}

\begin{table*}[htbp]
\caption{Zero-shot prediction results of LLMs.}
\label{tab: zero shot}
\footnotesize
\resizebox{\textwidth}{!}{
\begin{tabular}{c|ccc|ccc|ccc|ccc|ccc}
\hline
\multirow{2}{*}{LLM}                            & \multicolumn{3}{c|}{OFA} & \multicolumn{3}{c|}{GATGPT} & \multicolumn{3}{c|}{GCNGPT} & \multicolumn{3}{c|}{LLMAM2} & \multicolumn{3}{c}{ST-LLM}                        \\ \cline{2-16} 
                                                & MAE    & RMSE  & MAPE    & MAE     & RMSE   & MAPE     & MAE     & RMSE   & MAPE     & MAE     & RMSE   & MAPE     & MAE           & RMSE           & MAPE             \\ \hline
NYCTaxi Pick-up $\rightarrow$ CHBike Drop-off   & 3.57   & 5.72  & 59.26\% & 3.25    & 5.34   & 59.35\%  & 3.49    & 5.64   & 59.06\%  & 3.23    & 5.74   & 72.14\%  & \textbf{3.12} & \textbf{5.01}  & \textbf{55.12\%} \\
NYCTaxi Pick-up  $\rightarrow$ CHBike Pick-up   & 3.61   & 5.98  & 59.55\% & 3.29    & 5.60   & 59.71\%  & 3.53    & 5.91   & 59.14\%  & 3.25    & 5.15  & 88.52\%  & \textbf{3.06} & \textbf{5.40}  & \textbf{50.94\%} \\
NYCTaxi Pick-up  $\rightarrow$ NYCTaxi Drop-off & 9.99   & 20.22 & 75.14\% & 10.00   & 21.16  & 68.03\%  & 11.03   & 21.86  & 70.32\%  & 11.02   & 22.34   & 94.31\%  & \textbf{9.31} & \textbf{18.68} & \textbf{66.42\%} \\ \hline
NYCTaxi Drop-off  $\rightarrow$ CHBike Drop-off & 3.58   & 5.72  & 59.33\% & 3.19    & 4.99   & 76.75\%  & 3.35    & 5.19   & 69.36\%  & 3.29    & 4.99   & 80.87\%  & \textbf{3.09} & \textbf{4.65}  & \textbf{52.73\%} \\
NYCTaxi Drop-off $\rightarrow$ CHBike Pick-up   & 3.62   & 5.99  & 59.55\% & 3.26    & 5.27   & 79.41\%  & 3.43    & 5.49   & 71.76\%  & 3.33    & 5.32   & 82.60\%  & \textbf{3.02} & \textbf{5.18}  & \textbf{68.27\%} \\
NYCTaxi Drop-off  $\rightarrow$ NYCTaxi Pick-up & 10.04  & 17.72 & 88.10\% & 9.67    & 17.76  & 73.46\%  & 8.09    & 14.58  & 50.99\%  & 11.14   & 20.57  & 94.03\%  & \textbf{8.02} & \textbf{13.21} & \textbf{46.16\%} \\ \hline
\end{tabular}}
\end{table*}

\subsection{Inference Time Analysis}

\begin{figure}[t]
    \centering
    \subfigure[NYCTaxi Pick-up.]{
    \includegraphics[width=0.47\linewidth]{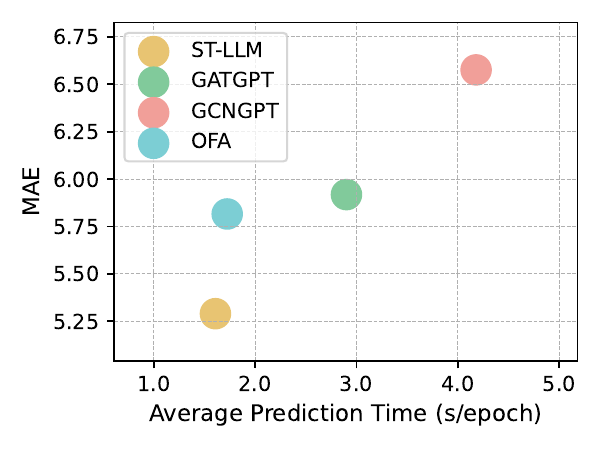}
    }\hfill
    \subfigure[CHBike Pick-up.]{
    \includegraphics[width=0.47\linewidth]{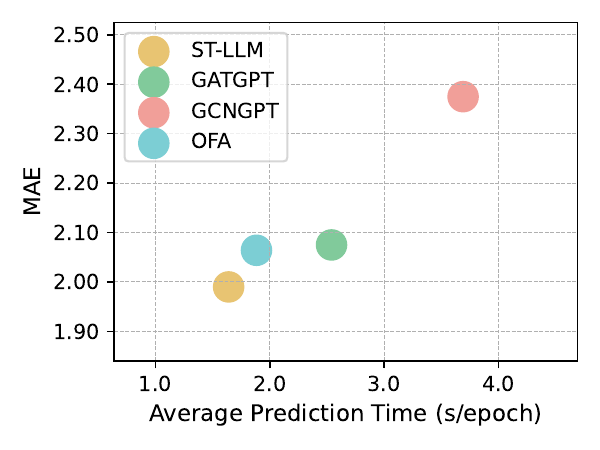}
    }\hfill
    \subfigure[NYCTaxi Drop-off.]{
    \includegraphics[width=0.47\linewidth]{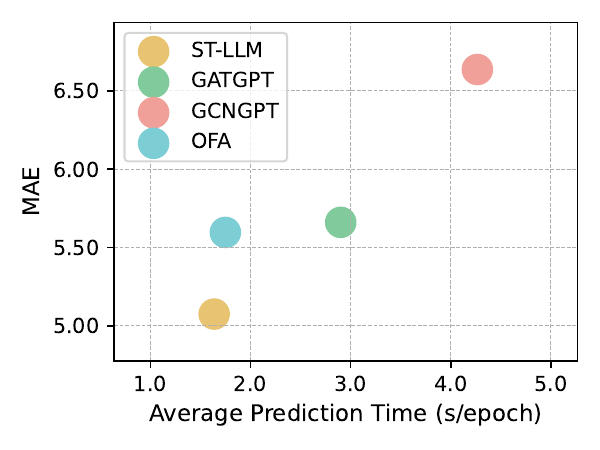}
    }\hfill
    \subfigure[CHBike Drop-off.]{
    \includegraphics[width=0.47\linewidth]{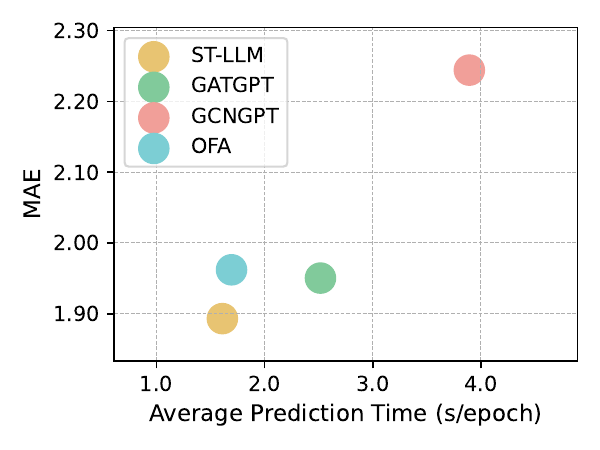}
    \hfill
    }
    \caption{Inference time of LLMs. 
    }
    \label{fig: time}
\end{figure}

Figure~\ref{fig: time} illustrates the trade-off between inference time and MAE for various LLMs on NYCTaxi and CHBike datasets. It's important to note from the outset that LLAMA2 is intentionally omitted from this comparative analysis because its inference time is significantly longer than that of the other LLMs under consideration, making it an outlier in terms of efficiency. For NYCTaxi datasets, ST-LLM achieves the lowest MAE while maintaining competitive inference times. Following closely is OFA, whose inference time is similar to ST-LLM, albeit with a slightly higher MAE. This suggests that spatial-temporal embedding and PFA do not slow down the inference speed of the LLM, yet enhance prediction accuracy. GATGPT and GCNGPT exhibit longer inference times and higher MAEs than ST-LLM, with GCNGPT being the slowest. This could be attributed to the fact that while GCN generally has a simpler structure, their combination with GPT might introduce extra computational complexity. This also implies that GAT's attention mechanism could be more efficient when combined with GPT. 

For CHBike datasets, we observe that the inference times and MAEs are generally lower than those for the NYCTaxi datasets, which might indicate differences in the datasets' complexity of the prediction tasks. For the CHBike Pick-up dataset, ST-LLM again achieves the lowest MAE. OFA follows with a closely competitive inference time but again falls slightly short on accuracy. GATGPT and GCNGPT show a consistent pattern, having longer inference times and higher MAEs, with GCNGPT being the slowest model. The CHBike Drop-off dataset reflects a similar pattern, which shows ST-LLM's robustness across different data scenarios. The OFA model remains a close second in speed, with a negligible increase in MAE. The trend of longer inference times for GATGPT and GCNGPT persists, with GCNGPT again taking the longest time among the models. In summary, ST-LLM stands out as the model providing the best balance between inference speed and predictive accuracy across both datasets. 

\subsection{Few-Shot Prediction}

In few-shot prediction, LLMs are trained with just 10\% of data. The experimental results are in Table~\ref{tab: few shot}. From the results, we can see that ST-LLM is superior in recognizing complex patterns from limited data, and we attribute this to the knowledge activation in our PFA LLM. While the LLAMA2 model presents competitive results, especially on the NYCTaxi datasets. However, it does not consistently surpass the performance of ST-LLM. For instance, on the NYCTaxi Pick-up dataset, ST-LLM achieves a noteworthy 7.06\% reduction in MAE compared to LLAMA2. The OFA, GATGPT, and GCNGPT, although commendable in their performances, do not match the superior results of ST-LLM. Notably, despite OFA's better performance on the CHBike Drop-off dataset, ST-LLM still outperforms it with a 9.15\% improvement in MAE. Compared with GATGPT and GCNGPT, ST-LLM shows remarkable average improvements of over 39.21\% and 7.80\% in MAE across all datasets, respectively. This significant difference highlights the robustness of ST-LLM in efficiently handling scenarios with limited data.

\subsection{Zero-Shot Prediction}

The zero-shot prediction experiments evaluate the intra-domain and inter-domain knowledge transfer capabilities of various LLMs. Each LLM in this evaluation predicts traffic flow in the CHBike dataset after being trained using only data from the NYCTaxi dataset, without prior exposure to the CHBike dataset. The results of zero-shot prediction are depicted in Table~\ref{tab: zero shot}. In terms of intra-domain transfer, such as predicting the NYCTaxi drop-off flow based on the NYCTaxi pick-up flow, ST-LLM demonstrates its ability to maintain high accuracy. The results also show that ST-LLM exhibits exceptional performance in inter-domain scenarios, such as transferring from NYCTaxi datasets to CHBike datasets. The ST-LLM consistently achieves the lowest error rates, indicating a robust ability to adapt to new domains without retraining. The results proved that the OFA is not a good zero-shot predictor. GATGPT and GCNGPT show competent adaptability but still fall short compared to ST-LLM's performance, particularly in challenging inter-domain transfers. LLAMA2 performs well in most inter-domain scenarios, and its performance is second only to ST-LLM. In conclusion, the zero-shot prediction results reinforce the adaptability and predictive strength of ST-LLM. We attribute this success to our PFA strategy being better at activating the LLM’s knowledge transfer and reasoning capabilities when performing traffic prediction tasks.

\section{Conclusion}
\label{sec: conclusion}

ST-LLM shows promise in adapting large language models for traffic prediction by embedding traffic data into spatial-temporal representations for LLMs. A partially frozen attention strategy is proposed to adapt the LLM to capture global spatial-temporal dependencies in traffic prediction. Our empirical studies show that the proposed ST-LLM performs better than the state-of-the-art traffic prediction models and LLMs. Future work will explore LLM for multi-task learning, such as incorporating traffic imputation, generation, and anomaly detection.

\section*{Acknowledgment}

This study is supported under the RIE2020 Industry Alignment Fund – Industry Collaboration Projects (IAF-ICP) Funding Initiative, as well as cash and in-kind contribution from the industry partner(s).

\bibliographystyle{IEEEtran}
\bibliography{ref}

\end{document}